\documentclass[journal]{IEEEtran}
\usepackage[utf8]{inputenc}
\usepackage{graphicx}
\usepackage{booktabs}
\usepackage{amsmath}

\usepackage{amsfonts}
\usepackage{multirow}

\usepackage{verbatim}

\usepackage{bbding}

\usepackage{color}
\usepackage{colortbl}  
\usepackage{xcolor}

\usepackage{graphicx}
\usepackage{textcomp}
\usepackage{float}

\usepackage{booktabs}
\usepackage{amsfonts}
\usepackage{amsthm,amsmath}
\usepackage{mathrsfs}
\usepackage{multirow}
\usepackage[noend]{algpseudocode}
\usepackage{algpseudocode}
\usepackage{diagbox}
\usepackage{color}
\usepackage{colortbl}  
\usepackage{xcolor}
\usepackage{array} 
\usepackage{bbding} 
\usepackage{graphicx}
\usepackage[marginal]{footmisc}

\begin{document}
	\title{Toward Realistic Camouflaged Object Detection: Benchmarks and Method}
	%
	%
	%
	
	\author{
			Zhimeng Xin,~\IEEEmembership{}
			Tianxu Wu,~\IEEEmembership{}	
			Shiming Chen,~\IEEEmembership{}
				Shuo Ye,~\IEEEmembership{}			
				Zijing Xie,~\IEEEmembership{}
				Yixiong Zou,~\IEEEmembership{}	
			 Xinge You,~\IEEEmembership{Senior Member,~IEEE,
			}
			and Yufei Guo~\IEEEmembership{}
		
		\thanks{		

			Z. Xin is with the School of Cyber Science and Engineering, Huazhong University of Science and Technology, Wuhan 430074, China (e-mail: zhimengxin@hust.edu.cn).

			T. Wu, S. Chen, S. Ye, and X. You are with the School of Electronic Information and Communications, Huazhong University of Science and Technology, Wuhan 430074, China (e-mail: wutianxu@hust.edu.cn; gchenshiming@gmail.com; shuoye@hust.eud.cn; youxg@mail.hust.edu.cn).
			
			Y. Zou is with the School of Computer Science \& Technology, Huazhong University of Science and Technology, Wuhan 430074, China (e-mail: yixiongz@hust.edu.cn). 
				
			Yufei Guo is with the Intelligent Science and Technology Academy of CASIC, Beijing 100041, China (e-mail: yufeiguo@pku.edu.cn).
		
			
		}
	}

\maketitle


\begin{abstract}

Camouflaged object detection (COD) primarily relies on semantic or instance segmentation methods. While these methods have made significant advancements in identifying the contours of camouflaged objects, they may be inefficient or cost-effective for tasks that only require the specific location of the object. Object detection algorithms offer an optimized solution for Realistic Camouflaged Object Detection (RCOD) in such cases.  However, detecting camouflaged objects remains a formidable challenge due to the high degree of similarity between the features of the objects and their backgrounds. Unlike segmentation methods that perform pixel-wise comparisons to differentiate between foreground and background, object detectors omit this analysis, further aggravating the challenge. To solve this problem, we propose a camouflage-aware feature refinement (CAFR) strategy. Since camouflaged objects are not rare categories, CAFR fully utilizes a clear perception of the current object within the prior knowledge of large models to assist detectors in deeply understanding the distinctions between background and foreground. Specifically, in CAFR, we introduce the Adaptive Gradient Propagation (AGP) module that fine-tunes all feature extractor layers in large detection models to fully refine class-specific features from camouflaged contexts. We then design the Sparse Feature Refinement (SFR) module that optimizes the transformer-based feature extractor to focus primarily on capturing class-specific features in camouflaged scenarios. To facilitate the assessment of RCOD tasks, we manually annotate the labels required for detection on three existing segmentation COD datasets, creating a new benchmark for RCOD tasks. Extensive experiments on our proposed datasets demonstrate that the CAFR strategy significantly improves the model's foreground and background recognition abilities and enhances RCOD task performance. Code and datasets are available at: 
https://github.com/zhimengXin/RCOD.

\end{abstract}

%
	

\section{Introduction}

\IEEEPARstart{C}{amouflaged} object detection (COD) currently relies primarily on semantic or instance segmentation methods \cite{COD1,COD4,COD6,segmar}. While such methods have significantly advanced in identifying the contours of camouflaged objects, they may be less efficient for tasks that only require the specific location of the object, e.g., search and rescue operations \cite{sso} or military strikes \cite{milistrike}. Furthermore, segmentation tasks often entail high annotation costs, particularly in instance segmentation which requires annotating the boundary coordinates of each object and determining its category. For low-cost tasks like camouflaged object counting \cite{obconting}, such annotation costs are undoubtedly an unnecessary expense. Object detection technology offers an optimized solution toward \textbf{R}ealistic \textbf{C}amouflaged \textbf{O}bject \textbf{D}etection (RCOD) in the aforementioned scenarios \cite{obconting2}. It only requires simple bounding boxes to precisely locate camouflaged objects and determine their categories, fulfilling task requirements while reducing annotation costs.

\begin{figure}[t]
	\begin{center}
		\includegraphics[width=0.47\textwidth]{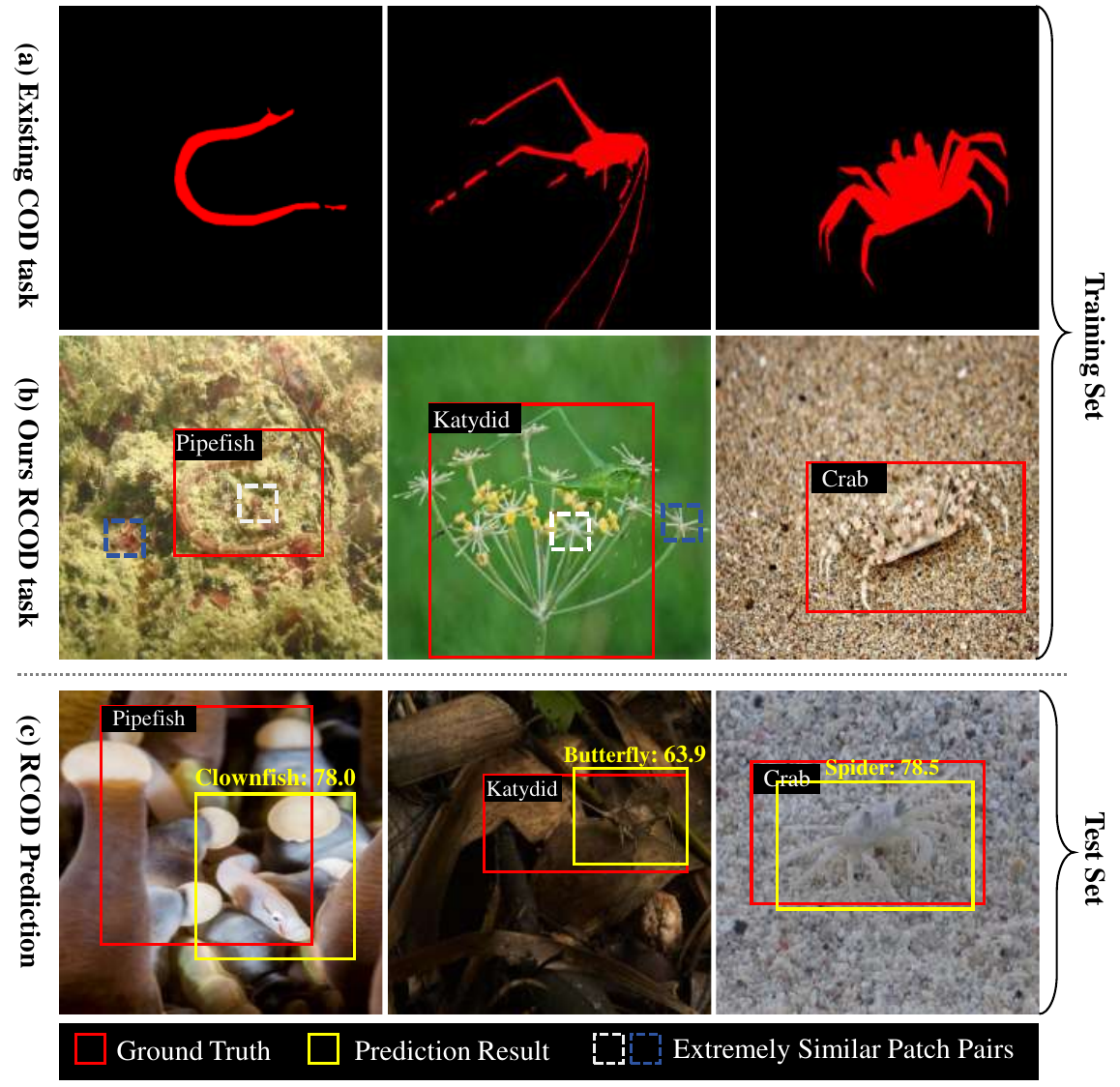}
	\end{center}
	\setlength{\abovecaptionskip}{-0.1cm} 
	\caption{Comparison of (a) existing COD and (b) our RCOD tasks and (c) visualization of the challenge in the RCOD task. As for RCOD prediction, we use the trained GLIP model \cite{glipv2} on our proposed dataset to visualize the detection results. Since the background and foreground are extremely similar, the direct application of large models pre-trained on mostly well-defined object contour scenes still leads to misidentification of camouflaged objects (c). Furthermore, the proposed datasets contain bounding boxes that encompass sparse category features, e.g., the bounding boxes for the classes \textit{Pipefish} and \textit{Katydid} contain less than half of the class-specific features (b). This situation further reduces GLIP's ability to detect these classes (c), because it employs the swin-transformer \cite {swin} as the feature extractor. This extractor focuses on the similarity relationships between pairs of patches, which can lead to confusion over camouflaged object features when assessing the similarities between blue-boxed and white-boxed patches (b).}
	\vspace{-0.3cm} 
	\label{motivation}
	\vspace{-0.2cm} 
\end{figure}

However, the currently available public datasets related to COD only provide mask labels for segmentation tasks and lack bounding boxes and corresponding class labels necessary for detection tasks. To address this limitation, we propose three new datasets to establish valuable benchmarks in the field of RCOD research. Specifically, we manually annotate bounding boxes and corresponding class labels, creating novel COD10K-D, NC4K-D, and CAMO-D based on existing COD datasets, including COD10K-v2 \cite{cod10k}, NC4K \cite{nc4k}, and CAMO \cite{camo}, suitable for the detection assessment.

Traditional object detectors are primarily categorized into the following groups: the one-stage \cite{yolov1,yolov7,yolov8} or two-stage-based frameworks \cite{fasterrcnn,cascade}, and transformer-based end-to-end architectures \cite{detr}, which rely on features such as textures, shapes, and colors for detection. Unfortunately, as for RCOD, the challenge arises from the high degree of similarity between the features of the object and those of the background \cite{preynet,zoom,fsel}. Unlike segmentation methods that perform pixel-wise comparisons to differentiate between foreground and background, such detectors omit this analysis in camouflaged scenarios. As a result, they face significant difficulties adapting to variations in object appearance and often lack the effective integration of contextual information, leading to poor performance in detecting these camouflaged objects.

Fortunately, recent advancements in large detection models \cite{glipv1,glipv2,Yolo-world,dino} that leverage large-scale data pre-training and robust semantically-aware feature extractors have significantly boosted the performance of traditional detectors. Since camouflaged objects are not rare categories, leveraging a clear understanding of an object within the prior knowledge of such large models to assist the detector in learning scenarios where this object is camouflaged. While these algorithms have shown strong adaptability in RCOD tasks, the direct application of large models pre-trained on mostly well-defined object contour scenes still leads to misidentification of camouflaged object features, as shown in Fig. \ref {motivation}(c). To tackle this challenge, a promising approach is to fine-tune these large models to better adapt to downstream tasks. Existing fine-tuning strategies typically involve freezing the feature extraction layers and retraining only the final fully connected layer, or introducing specific adaptation parameters in the final layer for the new task. Nevertheless, the important details, e.g., edge information or color variations of the camouflaged objects, are likely to have been significantly abstracted in the final layer of the feature extractor. Consequently, fine-tuning the final layer parameters makes model hard to effectively capture these easily confused features since the foreground and background of camouflaged objects are extremely similar.

Furthermore, some bounding boxes exhibit sparse category features within the proposed datasets. For instance, as illustrated in Fig. \ref{motivation}(b), the bounding boxes for the classes \textit{Pipefish} and \textit{Katydid} contain fewer than half of the class-specific features. This sparsity may hinder the performance of large models that rely on the transformer-based feature extractor \cite{swin} in accurately identifying these classes. The issue arises primarily because transformer-based feature extractors the similarity relationships between pairs of patches, which can lead to confusion over camouflaged object features when assessing the similarities between blue-boxed and white-boxed patches, as shown in Fig. \ref{motivation}(b).

Here, we explore how to alleviate the challenges associated with detecting camouflaged objects, building upon existing large detection models tailored for RCOD.
To achieve this goal, we consider updating a few parameters across all feature extraction layers of the  detection large model to ensure better coordination among them. This fine-tuning strategy allows the model to capture more crucial information for detection, including low-level features, e.g., edge contours and color variations, and high-level features, e.g., the overall shape of camouflaged objects and their semantic information. This approach enhances the model's ability to identify camouflaged objects that closely resemble their backgrounds through a deep exploration of the differences and connections between objects and backgrounds at multiple feature levels. 
Furthermore, freezing certain parameters helps prevent the model from overfitting to local patterns learned from the limited camouflaged data, while still effectively leveraging general knowledge. On the other hand, scaling down the size of the bounding boxes for large-scale camouflaged objects allows more class features to fit within an image patch, optimizing the performance of transformer-based feature extractors in the RCOD task.

Based on the analysis above, we propose a camouflage-aware feature refinement (CAFR) strategy that fully utilizes a clear perception of the current object within the prior knowledge of large models to assist detectors in deeply understanding the distinctions between background and foreground in RCOD tasks. Specifically, in CAFR, we first introduce the Adaptive Gradient Propagation (AGP) module, which fine-tunes all feature extractor layers of the model to fully refine class-specific features from camouflaged scenarios by restricting the proportion of gradient return from the feature extractor. We then design the Sparse Feature Refinement (SFR) module to reduce the size of camouflaged objects, which allows the model to focus on more class-specific features from the image’s local areas in sparse feature scenarios. Extensive experiments conducted on our proposed datasets demonstrate that the CAFR strategy significantly improves the model’s foreground and background recognition abilities and enhances RCOD task performance.

Our contributions can be summarized as follows:
\begin{itemize}
	\item 
	To the best of our knowledge, we are the first to transform camouflaged object segmentation into object detection tasks. To promote the sustainable development of detection tasks, we propose three new datasets to establish valuable benchmarks in the field of RCOD research.

	\item 
	
	We propose a CAFR strategy that fully utilizes a clear perception of the current object within the prior knowledge of large models to assist detectors in deeply understanding the distinctions between background and foreground in RCOD tasks.
	
	\item 
	In CAFR, we introduce the AGP module that fine-tunes all feature extractor layers in large models to fully refine class-specific features from camouflaged scenarios. We then design an SFR module that optimizes the transformer-based feature extractor to focus primarily on class-specific features in sparse feature scenarios.   
	
	\item 
	Extensive experiments conducted on our proposed datasets demonstrate that the CAFR strategy significantly improves the model’s foreground and background recognition abilities and enhances RCOD task performance.    
	
\end{itemize}

\section{Related Work}

\subsection{Camouflaged Object Detection} 

The goal of COD is to identify and locate objects that seamlessly blend with their background, making them hard to differentiate \cite{COD1,COD2,COD3,tip1,tip2,tip3}. To tackle this problem, current COD research primarily relies on algorithms based on semantic or instance segmentation \cite{COD4,COD5,COD6}. This is because segmentation algorithms excel in learning intricate features of objects and capturing more fine-grained information \cite{YE2023103837} by classifying each pixel, unlike object detection. Although these methods have made significant progress in accurately outlining boundaries of camouflaged objects, they might not be the most efficient for tasks that only require knowing the precise location of the object, such as search and rescue operations \cite{sso} or military strikes \cite{milistrike}. Moreover, the pixel-wise learning approach of segmentation algorithms significantly increases annotation costs, demanding expert-level annotations for uncommon classes. Object detection technology offers a more streamlined solution toward RCOD in the aforementioned scenarios. It only requires basic bounding boxes to precisely locate camouflaged objects and determine their categories, fulfilling task requirements while reducing annotation expenses. Therefore, object detection algorithms continue to be a valuable avenue for further research.

\subsection{Object Detection} 

Traditional object detectors are mainly categorized into the following groups: one-stage and two-stage frameworks, as well as end-to-end architectures based on transformer. One-stage detectors (such as early YOLO algorithms and SSD) \cite{yolov1,yolov2,ssd} perform object localization and classification simultaneously in a single step, hence offering faster speeds suitable for real-time applications. However, this approach may lack accuracy. Two-stage detectors (such as the R-CNN series) \cite{fasterrcnn} first generate candidate regions and then perform classification and regression on these regions, typically providing higher detection accuracy but with higher computational complexity and time costs. End-to-end architectures based on transformer (like DETR \cite{detr} and Deformable DETR \cite{deformabledetr}) have gradually emerged in recent years. These methods effectively handle global information through self-attention mechanisms to improve the preformance. However, the challenge of detecting camouflaged objects arises from the high degree of similarity between the features of the object and those of the background, which constrains the ability of such detection models to distinguish them accurately. 

Fortunately, as the field of object detection advances, the integration of multimodal detection algorithms based on image-text pairs has played a crucial role in improving detection performance. For example, The GLIP model \cite{glipv1,glipv2} learns the correspondence between language and images during the pre-training phase, enabling it to more accurately identify and locate objects in fine-tuning tasks. YOLO-World \cite{Yolo-world} combines the YOLO architecture with textual information, enhancing the understanding of semantic information in images and thereby improving detection accuracy. Grounding DINO \cite{dino} is a transformer-based object detection method that effectively handles global information and enhances detection accuracy through self-attention mechanisms. Moreover, BigDetection \cite{bigdetection} has also made significant progress driven by large-scale data pre-training. Nevertheless, these methods tend to rely heavily on the generic features of large models, which are generally derived from well-defined contour scenes. As a result, they struggle to adapt flexibly to the complex feature discrepancies between camouflaged objects and their backgrounds.

\subsection{Fine-tuning Large Models}

With the remarkable success of large-scale pre-trained models in areas such as natural language processing and computer vision, effectively fine-tuning these models for specific tasks has emerged as a critical research focus \cite{samadapter,ft1,ft2,ft3,ft4,ft5}. This provides a solution for extending the generalization of large models to COD tasks. Several effective fine-tuning strategies include adding adapters \cite{clipadapter,tipadapter,samadapter}, selective parameter tuning \cite{ft1,ft2,ft3,ft4,ft5}, etc. Adapters are lightweight modules designed specifically for fine-tuning tasks \cite{adapter1,adapter2}. The core idea involves introducing task-specific parameters by inserting adapter modules between specific layers without altering the original structure of the large model. For instance, Houlsby \textit{et al}. \cite{adapter1} proposed inserting small network modules between each layer or specific layers of a pre-trained model, enabling more efficient multi-task learning for large models. Pfeiffer \textit{et al.} \cite{adapter2} further refined this method with AdapterFusion, allowing the non-destructive combination of adapter modules for different tasks to enhance knowledge sharing between tasks. In the visual domain, CLIP-Adapter \cite{clipadapter} was introduced for fine-tuning lightweight residual feature adapters, albeit requiring additional training. Zhang \textit{et al.} \cite{vitadapter} proposed TIP-Adapter, which does not need extra training and retains the advantages of CLIP-Adapter. Furthermore, some adapters have been applied in detection or segmentation tasks, e.g., ViT-Adapter \cite{vitadapter} and SAM-Adapter \cite{samadapter}. Selective parameter tuning involves fine-tuning specific subsets of existing parameters to improve model performance for downstream tasks \cite{ft2,ft3,ft4,ft5}. However, these techniques predominantly address scenarios with clear visual contours. Our proposed CAFR tackles the challenge of extracting camouflaged objects that closely resemble their backgrounds by optimizing Transformer-based feature extractors and fine-tuning the entire feature extraction process.

\begin{figure*}[t]
	\begin{center}
		\includegraphics[width=1\textwidth]{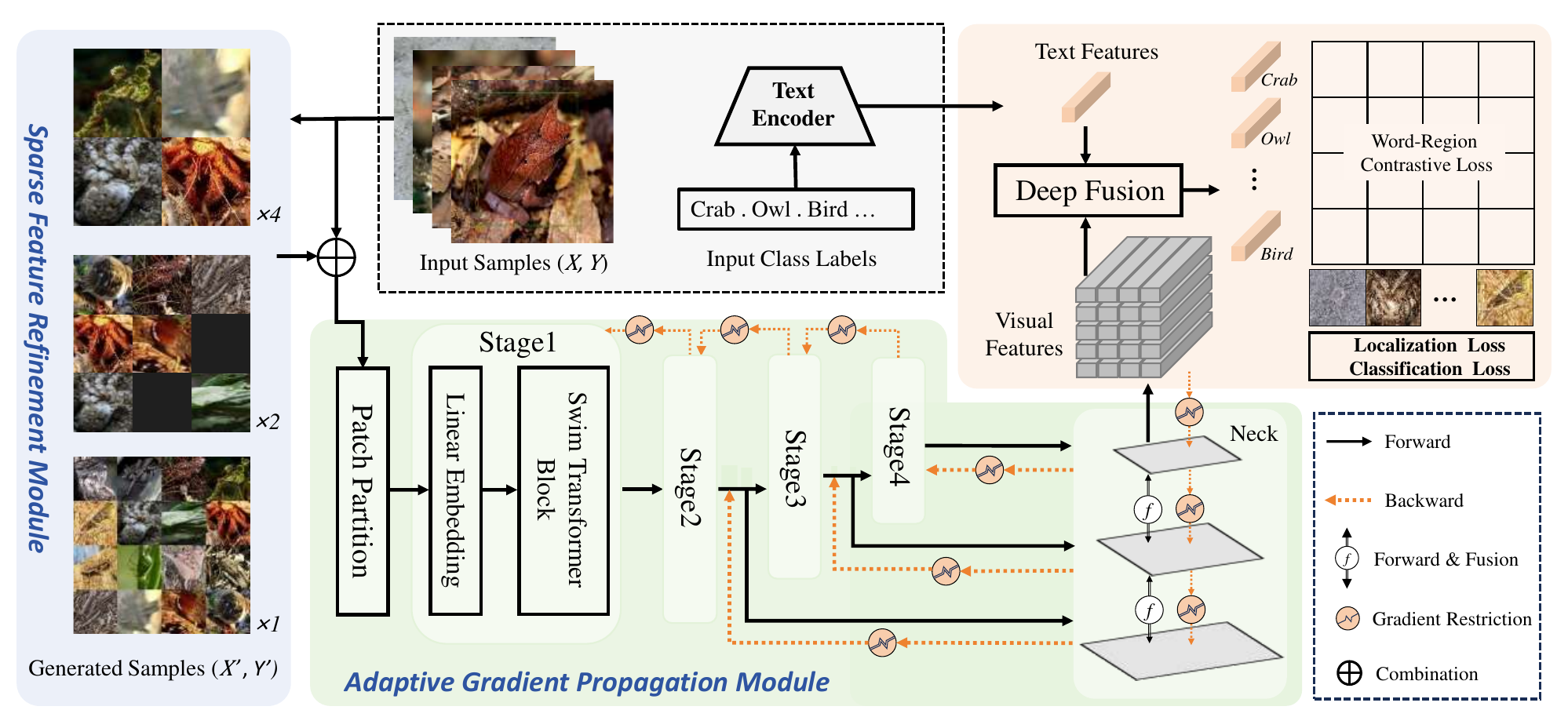}
	\end{center}
	\setlength{\abovecaptionskip}{-0.1cm} 
	\caption{The architecture of our proposed CAFR approach with SFR and AGP modules. Concerning the SFR module, within an input batch, each box is cropped to $W=200$ and $H=200$ from the input samples and randomly positioned on a square canvas. It is assumed that there are 16 boxes in a batch for showcasing the SFR module. In addition, $\times 1$ denotes obtaining 1 new sample, $\times 2$ denotes obtaining 2 new samples, and $\times 4$ denotes obtaining 4 new samples. As for the AGP module, it confines the backward pass of the detection head to the neck and backbone phases.}
	\label{method}
\end{figure*}

\section{Methodology}


In this section, we first introduce the task definition for our proposed RCOD task. We then illustrate the implementation details of our proposed CAFR approach that incorporates APG and SFR modules.

\section{Task Definition}

In this paper, given input data $\mathbb{D}$, RCOD aims to identify various objects in images and generate bounding boxes and their corresponding class labels for each object. Given an input sample $(X_i, Y_i) \in \mathbb{D}$, where $X_i$ represents the input image and its corresponding label  $Y_i=\left\{ y_n, b_n\right\}_{n=1}^{N}$. Here, $N$ represents the number of objects in an image. The bounding box $b_i = (x_{min}, y_{min}, x_{max}, y_{max})$ is typically defined using four parameters, where $(x_{min}, y_{min})$ denotes the coordinates of the top-left corner, and $(x_{max}, y_{max})$ indicates the coordinates of the bottom-right corner. Therefore, object detection in camouflaged scenarios can be defined as the following optimization problem

\begin{equation}
	\begin{aligned}
		\text{argmax}_{b, y} \sum_{n=1}^{N} \mathcal{L}(b_n, y_n, \mathbb{D})
		\label{L},
	\end{aligned}
\end{equation}
where $\mathcal{L}$ is the loss function used to evaluate the difference between the predicted bounding boxes and class labels.

\subsection{Why need Camouflage-Aware Feature Refinement in RCOD?}

The challenge of detecting camouflaged objects arises from the high degree of similarity between the features of the object and those of the background. Segmentation algorithms are a type of pixel-level classification technique that can effectively distinguish camouflaged objects from backgrounds by analyzing subtle differences at the pixel level, e.g., textures and variations in lighting. In contrast, detection algorithms often struggle to extract features within limited bounding boxes due to background interference, making it challenging to extract discriminative high-level features. Fortunately, the emergence of large detection  models with stronger representational abilities has reduced the gap between detection and segmentation algorithms in camouflaged scenarios.

To further bridge the gap between detection and segmentation algorithms in this scene, a promising but not the only approach is fine-tuning large detection models to better adapt to the RCOD task. Existing fine-tuning strategies \cite{adapter1,adapter2} typically involve freezing the feature extraction layers and retraining only the final fully connected layer, or inserting an adapter in the last layer of the feature extractor for learning the new task.
However, the important details, e.g., edge information or color variations of the camouflaged objects, are likely to have been significantly abstracted in the final layer of the feature extractor. Consequently, fine-tuning the final layer parameters makes model hard to effectively capture these easily confused features since the foreground and background of camouflaged objects are extremely similar.
To handle this problem, we propose an appropriate CAFR strategy that incorporates AGP and SFR, aimed at enhancing the detection large model's ability to distinguish between foreground and background in the RCOD task.

\subsection{Adaptive Gradient Propagation Module}

To refine the learning of object and background features, inspired by the decoupling concept \cite{defrcn}, we propose the AGP module to effectively fine-tune the large model in adapting the RCOD task. Our approach is based on image-text pair detection models \cite{glipv1,glipv2,dino}, where we utilize a one-stage detector \cite{yolov1,yolov2} to improve the performance of the RCOD task by introducing the proposed fine-tuning strategy. Following the one-stage detector, the detection loss can be given by

\begin{equation}
	\begin{aligned}
		\mathcal{L} = \mathcal{L}_{\text{bbox}} + \mathcal{L}_c + \mathcal{L}_{\text{cls}},
		\label{L}
	\end{aligned}
\end{equation}
where $\mathcal{L}_{\text{bbox}}$ is the localization loss that helps the model estimate the quality of the predicted bounding box in terms of its localization, $\mathcal{L}_c$ represents contrastive loss, and $\mathcal{L}_{\text{cls}}$ is classification loss.

Regarding the AGP module, as illustrated in Fig. \ref{method}, it restricts the backward pass of the detection head to the feature extraction phase. Simultaneously, it constrains the backward pass from the neck to the backbone in a similar fashion during the feature extraction process. Through this approach, AGP preserves the original parameter structure of the pre-trained model, making minimal parameter adjustments tailored to specific tasks. This ensures that fine-tuning the model for the COD task does not result in overfitting.

Denote $\theta_b$, $\theta_n$, $\theta_h$ as the parameters of the backbone, neck, and detection head of the large model, and denote $\theta$ as the parameters obtained from one training, i.e., $\theta = \left\{ \theta_b, \theta_n, \theta_h \right\}$  After one gradient-limited backward pass, the parameters $\theta = \left\{ \theta_b, \theta_n, \theta_h \right\}$ can be given by

\begin{equation}
	\begin{aligned}
		\theta_h &= \theta_h - \lambda \left( \alpha \frac{\partial \mathcal{L}}{\partial \theta_h} \right) \\
		\theta_n &= \theta_n - \lambda \left(\alpha \frac{\partial \mathcal{L}}{\partial \theta_n} \right) \\
		\theta_b &= \theta_b - \lambda \left(\alpha \frac{\partial \mathcal{L}}{\partial \theta_b} \right),
	\end{aligned}
\end{equation}
where $\alpha$ is the learning rate and $\lambda$ is the restriction factor to control the size of parameters that can participate in training.
By substituting the above formula into Eq \ref{L}, the AGP-based loss can be expressed as

\begin{equation}
	\begin{aligned}
		\mathcal{L} = \mathcal{L}_{head}(M_h(M_n(M_b(X_i; \theta_b);\theta_n); \theta_h),Y)),
	\end{aligned}
\end{equation}
where $M(\cdot)$ represents learnable function.
In RCOD, we adopt \cite{conloss} as our contrastive loss $\mathcal{L}_c $, GIoU \cite{giou} as our Localization loss $\mathcal{L}_{\text{bbox}}$, and focal loss \cite{focalloss} as our classification loss $\mathcal{L}_{\text{cls}}$.

\subsection{Sparse Feature Refinement Module} 



The current large models for object detection utilize the swin-transformer \cite{swin} as the feature extractor and FPN \cite{fpn} for feature refinement.  These models incorporate self-attention-based feature extractors that are designed under the assumption that the outlines of instances are easily distinguishable \cite{selfattention}. This enables the models to capture long-range correlation by calculating similarities between image patches. However, in the case of camouflage scenarios, some bounding boxes encompass sparse category features, capturing less than half of the class-specific features, which may hinder the ability of models based on self-attention mechanisms to learn distinguishing features.

To optimize the efficiency of the self-attention mechanism, we propose the SFR module, which serves to reduce the size of the camouflaged object within an image. Through this approach, SFR ensures that each image patch primarily focuses on the most significant parts of the object region, and then selects different camouflaged objects with irrelevant information in other patches. Here, we introduce two implementations of SFR: one utilizing an online approach and the other employing an offline approach.

\subsubsection{Online Implementation of SFR}

To achieve multi-scale learning of objects in RCOD scenarios, we design SFR by randomly assembling all bounding boxes in a batch of images and treating each box as a \textit{pseudo-region} to scale objects. In each training iteration, we start by creating a large canvas that is evenly divided into square regions, such as a $2 \times 2$, $3 \times 3$, or $4 \times 4$ grid. We then randomly extract box-label pairs from the current batch, resizing all boxes to fill a square canvas while resetting the labels. For example, if there are 16 boxes in a batch, these boxes can be assembled into one canvas as a $4 \times 4$ pseudo-image, four canvases as four  $2 \times 2$ pseudo-images, or two $3 \times 3$ pseudo-images. Since two $3 \times 3$ require 18 boxes but the current batch cannot provide an extra two regions, black pixels are used as a background and randomly combined into the final canvas.  This enables the generation of multi-scale camouflaged objects without the need for manual labeling. The size of the assembled canvas remains fixed, and any extra parts are filled with black pixels.

Denoting the obtained pseudo samples in a batch as $(X', Y')$, they are trained together with the original image to train the detector. The training samples $(X, Y)$ can be represented as 

\begin{equation}
	\begin{aligned}
		(X, Y) = (X' + X, Y' + Y),	
	\end{aligned}
\end{equation}
where $(X, Y)$ represents the input sample $X$ along with its corresponding boxes and labels $Y$. This approach allows us to establish precise mappings between pseudo-regions of different scales in the assembled images and their corresponding labels, saving costs compared to re-labeling the dataset.

The online approach offers the advantage of directly processing input data to obtain scaled-down data, providing greater flexibility and convenience. However, due to the new images generated in a batch, this approach requires additional GPU usage.

\begin{figure}[t]
	\begin{center}
		\includegraphics[width=0.45\textwidth]{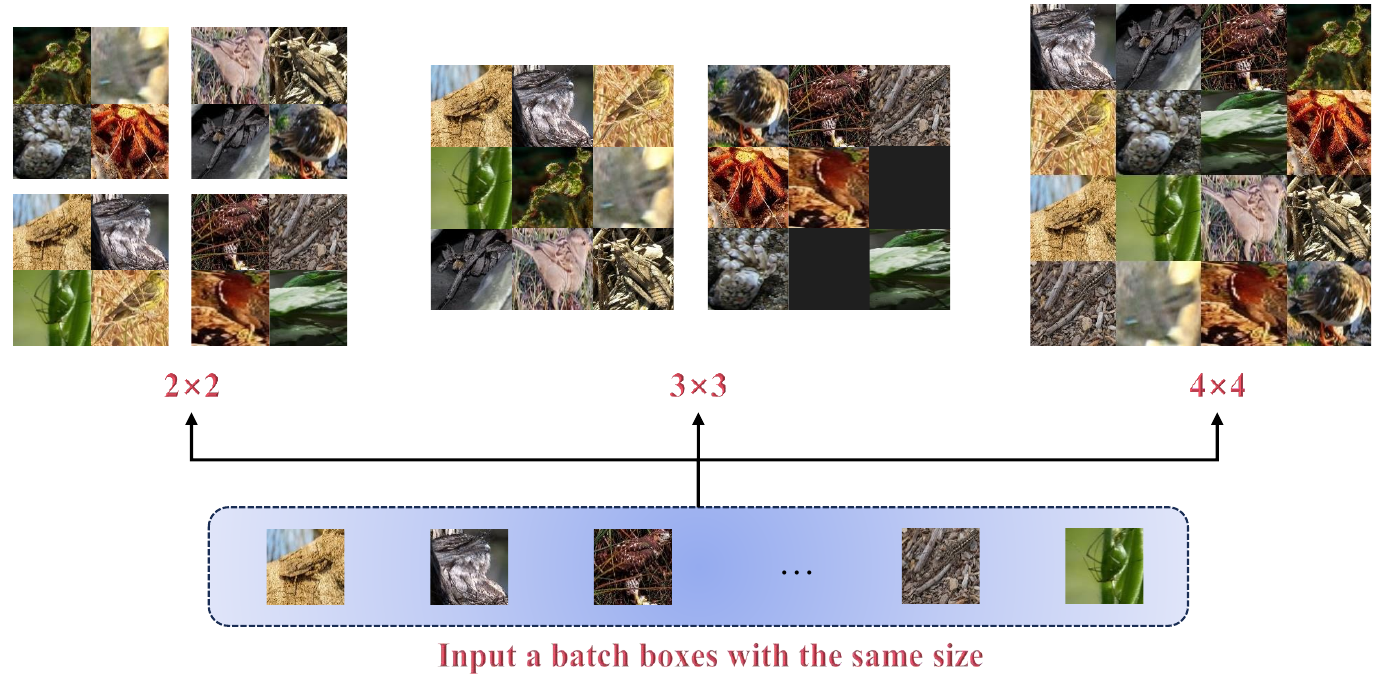}
	\end{center}
	\setlength{\abovecaptionskip}{-0.1cm} 
	\caption{SFR offline setting. In the SFR module, each box is cropped to $W=200$ and $H=200$ from the selected 16 bounding boxes and randomly positioned on a square canvas.}
	\label{sroffline}
	\vspace{-0.2cm} 
\end{figure}

\begin{figure*}[t]
	\begin{center}
		\includegraphics[width=0.9\textwidth]{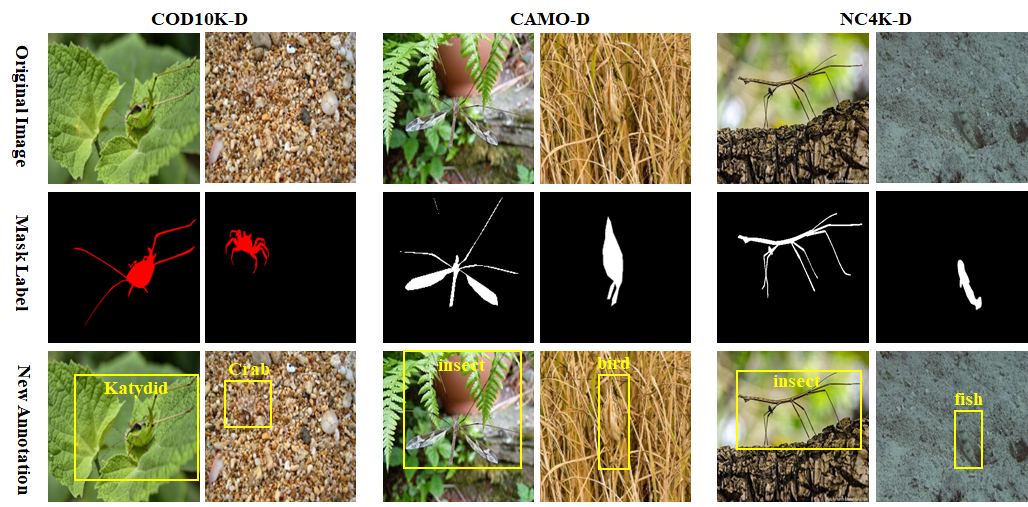} 
	\end{center}
	\setlength{\abovecaptionskip}{-0.1cm} 
	\caption{Comparison of annotations: existing vs. our annotations. Introducing these three new annotations establishes valuable benchmarks in the field of RCOD research.}
	\vspace{-0.3cm} 
	\label{data-image}
	\vspace{-0.2cm} 
\end{figure*}
\subsubsection{Offline Implementation of SFR}
The offline approach eliminates the need for additional CUDA memory usage, but requires manual pre-processing of data befor training.
In the offline process, we assume that a batch includes 16 bounding boxes. With this assumption in mind, we gather all the boxes from the entire dataset and randomly shuffle them. We then sequentially select 16 boxes at a time for canvas filling with new label settings. The filling method employed in this offline approach is the same as the one described in the main text for the online approach. 
Based on this fact, new samples of sizes $2 \times 2$, $3 \times 3$, and $4 \times 4$ are obtained simultaneously and showcased in Figure \ref{sroffline}. These new samples, acquired through the offline process, are subsequently merged with the original images to form a new dataset for training the detection network.

Offline approach requires manual pre-processing of data befor training, making it less flexible compared to the online method.  The experimental results in Section \ref{online} show that they have negligible performance differences. Therefore, the choice between the online and offline methods can be arranged reasonably based on the specific requirements of RCOD tasks.


Moreover, expanding instances at multiple scales for camouflaged object datasets requires a large number of region-level bounding boxes for annotation, which is both expensive and time-consuming in the annotation process. On the other hand, certain types of data may be rare, posing challenges in terms of acquiring additional training data for these models.Regardless of whether the implementation is online or offline, our SFR module addresses these issues by generating synthetic data.

\section{Novel Datasets for Benchmark}

\begin{table}[t]
	\centering	
	\tabcolsep=0.25cm
	\renewcommand{\arraystretch}{1.2}	
	\caption{Quantitative statistics on our proposed datasets. We present the counts of categories as well as the numbers of training and test images for our proposed datasets.}
	\begin{tabular}{l|ccc}
		\midrule
		
		Datasets  & Categories  & Training Images &  Test Images  \\ \midrule

		COD10K-D  & 68  & 6000  &  4000  \\ 
		NC4K-D  & 37 & 2863  &  1227  \\

		CAMO-D  & 43 & 744 & 497  \\

		\bottomrule
		
	\end{tabular}
	
	\label{datas}
\end{table}


The currently available public datasets related to COD primarily focus on camouflaged object segmentation. However, these datasets only provide mask labels for segmentation tasks and lack the bounding boxes and class labels crucial for object detection tasks. To overcome this limitation, we utilize the \textit{LabelMe} \footnote{https://github.com/wkentaro/labelme} tool to manually annotate labels for the creation of the COD10K-D, NC4K-D, and CAMO-D datasets, which are based on COD10K-v2 \cite{cod10k}, NC4K \cite{nc4k}, and CAMO \cite{camo}, respectively. Fig. \ref{data-image} illustrates the differences between the new annotations and the existing annotations in these three datasets.

In terms of annotation, we use mask labels to assist us in completing annotations. To generate bounding boxes, we begin by scanning the non-black pixels in the mask label to identify the coordinates of the furthest left, right, top, and bottom contiguous pixels, thereby creating rectangular boxes. However, relying exclusively on these bounding boxes can be problematic, as occlusions may disrupt the continuity of the object pixels. For example, as shown in Fig. \ref{data-image} from the CAMO-D dataset, a single object may have its contiguous pixels split into two parts, causing the algorithm to erroneously produce two separate boxes. Therefore, even with manually assigned annotations, it is necessary to remove or correct any inaccurate boxes. Fortunately, for class annotations, the COD10K-v2 dataset offers comprehensive category annotations, allowing us to select 68 valid classes for our detection task. Class annotations for other datasets are provided manually.

Based on the aforementioned annotation scheme, we follow the setup of the original dataset COD10K-v2, using 6000 images for training and 4000 images for testing, with 68 annotated class labels. Moving on to NC4K-D, we carefully select 4121 clearer images from the NC4K dataset \cite{nc4k} and annotate them with bounding boxes and 37 category labels. We then split this annotated dataset into a training set of 2863 images and a test set of 1227 images. Regarding the CAMO-D dataset, we combine 1000 training images and 250 test images from the original camouflaged CAMO dataset \cite{camo}. We diligently remove any controversial images and then annotate 1241 images for object detection with 43 diverse classes. From this annotated dataset, we randomly select 744 images for training and reserve 497 images for testing. Additionally, in Fig. \ref{datas}, we list the counts of categories as well as the numbers of training and test images for our proposed datasets.

By enriching existing COD datasets with boxes and class labels, our newly created COD10K-D, NC4K-D, and CAMO-D datasets offer valuable resources for realistic camouflaged detection tasks. Researchers can leverage such datasets to develop and evaluate robust object detection algorithms capable of addressing the challenges posed by camouflaged objects.

\section{Experiments on RCOD Detection}


In this Section, we first introduce the experimental settings of our proposed CAFR. We then conduct a series of comparative experiments to evaluate the effectiveness of CAFR in terms of detection performance on our proposed datasets. Finally, we investigate the impact of the components between SFR and APG, as well as the influence of different parameter settings on their performance.

\begin{table}[t]
	\centering	
	\tabcolsep=0.15cm
	\renewcommand{\arraystretch}{1.2}	
	\caption{Inference cost between segmentation and detection.  Vit: Vision Transformer, Swin: Swin Transformer, Par: Parameter, and GDino: Grounding Dino, T: tiny model, L: large model.}
	\begin{tabular}{l|c|c|c}
		\midrule
		
		Methods  & Backbone & Par (M) &   FLOPs (G) $\downarrow$ \\ \midrule
		
		\textit{Traditional Method} \\
		
		Mask RCNN \cite{maskrcnn} & ResNet50  & 41  &  289.97  \\ 
		Faster RCNN \cite{fasterrcnn} & ResNet50 & 34  &  \textbf{176.54}  \\ \midrule

		\textit{Large Model}  \\
		SAM \cite{sam} & Vit-T & 89 & 371.98  \\
		GDino + CAFR & Swin-T & 165 & \textbf{353.98 } \\  \midrule
		
		SAM \cite{sam} & Vit-L & 298 & 1314.89  \\
		GDino + CAFR  & Swin-L & 222 &  \textbf{524.68 } \\

		\bottomrule
		
	\end{tabular}
	
	\label{flops}
\end{table}

\subsection{Experimental Settings}

Our code is deployed under the MMDetection \cite{mmdetection} framework. For a fair comparison, all detection models presented in our experiments are reported by the MMDetection framework. Large detection models are utilized as our baseline. To optimize our model, we employ the AdamW optimizer with an initial learning rate of 0.0003 and a momentum of 0.05. Due to the limited number of samples for each class of RCOD data, we train 10 groups of random seeds for each model and present the average performance as our final results for robust evaluation of detection performance. Additionally, we assess the accuracy of our models using coco-style \cite{coco} mAP (IoU thresholds of .50:.05:.95), AP50 (IoU threshold of .50), AP75 (threshold of .75), APm (medium-sized objects), and APl (large-sized objects) indicators \cite{coco}, which provide an average measure of accuracy under various IoU thresholds for overlapping associations.

\begin{table}[t]
	\centering	
	\tabcolsep=0.25cm
	\renewcommand{\arraystretch}{1.2}	
	\caption{Detection versus segmentation in the camouflaged scene. 
	}
	\begin{tabular}{l|c|c}
		\midrule
		
		Methods   &  Backbone  & $S_{\alpha}$ / mIoU $\uparrow$  \\ \midrule
		\textit{Segmentation} \\
		SINet   \cite{cod10k} & ResNet50 &   77.0 \\
		SegMaR \cite{segmar}  & ResNet50 &  81.1 \\
		PreyNet  \cite{preynet} & ResNet50 &   81.0  \\    
		ZoomNet \cite{zoom}  & ResNet50 &  83.5  \\ 
		FSEL    \cite{fsel}  & ResNet50 &  87.3  \\  \midrule
		\textit{Detection} \\
		Faster R-CNN & ResNet50  & 95.7  \\
		Faster R-CNN & ResNet101  &  96.1  \\
		Faster R-CNN & Swin-L  &  \textbf{99.1}  \\
		
		\bottomrule
		
	\end{tabular}
	\label{miou}
	
\end{table}

\begin{table*}[t]
	\centering	
	\tabcolsep=0.07cm
	\renewcommand{\arraystretch}{1.2}	
	\caption{Preformance of various detection methods on COD10K-D, NC4K-D, and CAMO-D datasets. Here, Def-DETR is Deformable DETR \cite{deformabledetr}, Swin represents swin-trasformer with its tiny base, and large pre-trained models \cite{swin}, R50 and R101 are ResNet50 and RseNet101, respectively \cite{resnet}, GLIP with its tiny and large pre-trained models \cite{glipv2}, and GDino is Grounding Dion its tiny and base pre-trained models \cite{dino}. Bold font indicates the best result in the group. }
	\scalebox{1}{
		\begin{tabular}{l|c|ccccc|ccccc|ccccc}
			\midrule
			
			\multirow{2}{*}{Methods}  & \multirow{2}{*}{Backbone} & \multicolumn{5}{c|}{COD10K-D}   & \multicolumn{5}{c|}{NC4K-D}  & \multicolumn{5}{c}{CAMO-D}                                                                                       \\  
			
			& & mAP & AP50 & AP75 & APm & APl & mAP & AP50 & AP75 & APm & APl & mAP & AP50 & AP75 & APm & APl\\ \midrule
			
			\textit{Generic Methods} \\
			
			YOLOv7-L \cite{yolov7} & CSPDarknet & 3.8&8.2&2.8&1.0&4.0  & 6.8&14.2&6.0&1.7 &7.3&  5.4 & 10.2 & 5.5 & 8.5 & 5.4 \\
			
			Faster RCNN \cite{fasterrcnn} & RseNet50 & 8.3  &  21.0   & 5.0  & 4.8  &  8.8 & 19.2  &  39.7   & 16.0  & 8.1  &  20.2 & 4.7  &  12.4   & 2.2  & 3.5  &  5.3 \\
			
			YOLOv8-L \cite{yolov8} & CSPVOVNet & 9.7&16.8 &9.4 &2.6 &10.4 &  23.5 & 34.9 & 25.2 & 10.6 & 24.7  & 25.4 &  37.2 & 26.1 & 14.4 &26.7 \\

			Faster RCNN \cite{fasterrcnn} & RseNet101 & 10.8  &  24.4   & 7.7  & 9.2  &  11.6  & 23.0 &  47.2   & 20.1  & 10.4  &  24.0 & 9.3  &  21.1   & 6.9  & 9.9  &  10.1  \\
			
			Def-DETR \cite{deformabledetr} & RseNet50 & 12.2 & 23.1 &  11.4 & 6.5  & 13.1 & 27.4 &  49.6 & 27.9 & 14.0  & 29.7 & 13.3 & 26.9  & 12.4 & 9.7  &  14.1\\
			
			Def-DETR \cite{deformabledetr} & RseNet101 & 13.5&23.7&13.5&9.2&14.6&  30.9&54.4&32.0&12.4&32.5  &13.7&27.4&12.7&13.2&15.5 \\
			
			Cascade RCNN \cite{cascade} & RseNet101 & 15.3&27.4&15.9&8.5&16.4& 27.5 & 46.9 & 28.9 & 11.3 & 29.2 & 14.0 & 26.6 & 13.1 & 13.1 & 14.8\\
			
			Faster RCNN \cite{fasterrcnn}  & Swin-T  & 16.3 & 35.3 & 13.1 & 8.6  & 17.4 & 29.1 & 58.8 & 25.6 & 16.8  & 30.4 & 11.3 & 32.3 & 5.5 & 8.6  & 12.0 \\
			
			Faster RCNN \cite{fasterrcnn} & Swin-L  & 32.1 & 54.6 & 33.1 & 17.1  & 33.9  & 49.1 & 75.8 & 55.1 & 22.7  & 51.3  & 34.2 & 67.4 & 30.2 & 24.0  & 36.1 \\
			
			\midrule

			\textit{Large Models} \\
				
			GLIP \cite{glipv2} & Swin-T & 26.4  & 36.3  & 28.5  & 14.7   & 28.0  &  49.6  & 63.7  & 53.4  & 23.9   & 51.6 &  32.6 & 42.9  & 33.6  & 40.9   & 35.1 \\
			
			GLIP + CAFR & Swin-T &  28.8$^{\color{red}\uparrow\text{\textbf{2.4}}}$ & 38.2 & 31.0 & 16.4 & 30.6$^{\color{red}\uparrow\text{\textbf{2.6}}}$ &  51.3$^{\color{red}\uparrow\text{\textbf{1.7}}}$& 66.7 & 53.9 & 25.9  & 53.4$^{\color{red}\uparrow\text{\textbf{1.8}}}$  &  32.9 $^{\color{red}\uparrow\text{\textbf{0.3}}}$ & 42.2  & 32.8  & 35.9   & 36.0$^{\color{red}\uparrow\text{\textbf{0.9}}}$ \\
			
			GLIP \cite{glipv2} & Swin-L &  40.2 & 47.9 & 43.5 & 24.7 & 42.3 &  76.9 & 86.9 & 80.9 & 50.4 & 78.5  &  63.0 & 74.4 & 68.1 & 52.4 & 66.8 \\
			
			GLIP + CAFR  & Swin-L &  40.9$^{\color{red}\uparrow\text{\textbf{0.6}}}$  & 48.9  & 43.8  & 26.7   & 42.9$^{\color{red}\uparrow\text{\textbf{0.6}}}$  & 77.4 $^{\color{red}\uparrow\text{\textbf{0.5}}}$ &  88.8   &  81.8  & 52.6  &  79.4$^{\color{red}\uparrow\text{\textbf{0.9}}}$  & 63.5$^{\color{red}\uparrow\text{\textbf{0.5}}}$ & 75.1 & 68.3 & 48.7 &  67.6$^{\color{red}\uparrow\text{\textbf{0.8}}}$\\	
			
			GDino \cite{dino} & Swin-T &  44.8  & 56.0  & 47.9  & 23.5   & 47.8  &  69.8  & 81.0  & 72.1  & 37.5   & 72.4 &  48.0  & 59.1  & 52.4  & 40.7   & 52.2\\

			GDino + CAFR & Swin-T & 46.5$^{\color{red}\uparrow\text{\textbf{1.7}}}$ &58.6 &50.4 &26.9 &49.5$^{\color{red}\uparrow\text{\textbf{1.7}}}$ &   70.5$^{\color{red}\uparrow\text{\textbf{0.7}}}$ & 82.4 & 73.4 &33.7 &73.4$^{\color{red}\uparrow\text{\textbf{1.0}}}$ &  49.0$^{\color{red}\uparrow\text{\textbf{1.0}}}$ & 59.7 & 52.3 &43.7 &53.7$^{\color{red}\uparrow\text{\textbf{1.5}}}$\\

			GDino \cite{dino} & Swin-B &  58.7  & 70.9  & 63.1  & 23.6   & 62.3 &  79.9  & 90.5  & 84.6  & 54.8   & 81.5 &  68.6  & 80.6  & 75.1  & 55.7   & 73.0  \\ 
			
			GDino + CAFR & Swin-B & \textbf{60.6}$^{\color{red}\uparrow\text{\textbf{1.9}}}$ &\textbf{73.3} & \textbf{65.3} &\textbf{33.4} &\textbf{64.3}$^{\color{red}\uparrow\text{\textbf{1.0}}}$ & \textbf{81.2}$^{\color{red}\uparrow\text{\textbf{1.3}}}$ & \textbf{91.8} & \textbf{86.2} & \textbf{56.8} & \textbf{82.7}$^{\color{red}\uparrow\text{\textbf{1.2}}}$  & \textbf{70.0}$^{\color{red}\uparrow\text{\textbf{1.4}}}$ & \textbf{82.2} & \textbf{76.0} & \textbf{56.8} & \textbf{74.5}$^{\color{red}\uparrow\text{\textbf{1.5}}}$\\		
			\bottomrule	
			
	\end{tabular}}

	\label{datasets}	
\end{table*}

\subsection{Comparison of Detection and Segmentation on COD Tasks}

\textbf{Efficiency Verification and Analysis.}
Existing camouflaged object detection (COD) methods rely primarily on segmentation algorithms \cite{COD1,COD4,COD6}. While such methods have significantly advanced in identifying the contours of camouflaged objects, they may be less efficient for tasks that only require the specific location of the object. To verify this point, we present the results of detection and segmentation measured in floating point operations (FLOPs) in Table \ref{flops}. For a fair comparison, we select detection (Faster R-CNN) and segmentation (Mask R-CNN) frameworks that are structurally similar and utilize the same training images for both the COD10K and the proposed COD10K-D datasets. As shown in the table, Faster R-CNN demonstrates lower FLOPs compared to Mask R-CNN when using the same feature extractor. In addition, Mask R-CNN enhances Faster R-CNN by incorporating an additional branch specifically designed for object mask prediction, resulting in a higher parameter count.

By leveraging large models, we employ the Segment Anything Model (SAM) and Grounding Dino (GDino) for segmentation and detection tasks. The evaluation criteria are consistent with traditional methods. As shown in Table \ref{flops}, the results indicate that GDino significantly outperforms SAM in the COD task. Interestingly, from Table \ref{flops}, we found that despite SAM and GDino employing different backbones, the parameter count of SAM's tiny model is only half that of GDino's. Nevertheless, GDino still demonstrates a faster computation speed than SAM.

Based on the experimental results presented, we can conclude that segmentation requires the image encoder to analyze the entire image for each sample during the inference stage, leading to higher computational complexity. Furthermore, the mask decoder computes the value of each pixel in the image to generate the segmentation mask. In contrast, the inference process in detection focuses on identifying objects within the image and generating bounding boxes. Its computational load primarily lies in the extraction and analysis of image features by the feature extraction network. While computing image features is necessary, it does not involve complex per-pixel mask generation calculations. Determining the position of bounding boxes mainly requires calculating target coordinates based on the results of feature matching, with relatively lower computational intensity.

\textbf{Performance Verification and Analysis.}
To ensure fairness and align with existing COD methods, we focus on traditional detection models, assessing their ability to locate masks while intentionally excluding classification. Additionally, we implement a stricter set of mIoU thresholds of .50:.05:.95 for comparison against COD's $S_{\alpha}$. The experimental results are summarized in Table \ref{miou}. Our tests on the localization performance of COD reveal that the classical Faster R-CNN with a ResNet50 backbone achieves a mIoU of 95.7\%. Further improving upon this, the ResNet101 model reaches 96.1\%, while the Swin transformer model impressively achieves 99.1\%.

The results indicate that the primary challenge in object detection is not the inability to locate targets but rather the tendency for classification errors. Additionally, it is meaningless to compare proposal regions from object detection with segmentation masks. Proposal regions annotate all features considered potential targets, including clearly defined backgrounds. For example, in the second row of Figure 1, there is a Katydid insect positioned behind a plant. Our objective is to identify the insect concealed by the plant. If the insect is not labeled during training and only the bounding box is taken into account, the model will instinctively focus on learning the well-defined edges of the plant, leading to accurate bounding box predictions. In contrast, segmentation tasks cannot function in this manner. Therefore, comparing object detection tasks with segmentation tasks without considering the target category is inherently unfair to the task of object segmentation.

\subsection{Comparison Experiment on RCOD Tasks}

\textbf{Results on COD10K-D.}
We initially present the results obtained from traditional methods, including YOLO-based one-stage detectors \cite{yolov7,yolov8}, R-CNN-based two-stage detectors \cite{fasterrcnn,cascade}, and transformer-based end-to-end architectures \cite{deformabledetr}, as illustrated in Table \ref{datasets}. The table indicates that these models face significant challenges in accurately recognizing camouflaged objects, often relying on incorrect assumptions, especially evident in YOLOv7 \cite{yolov7} and the Faster R-CNN \cite{fasterrcnn} with the ResNet50 \cite{resnet} backbone.
 
 Fortunately, the use of large detection models \cite{glipv1,glipv2,dino} significantly improves the accuracy to over 26\%, providing a better understanding of the disguised object rather than mere speculation. This highlights the capability of well-trained pre-training models to bridge the cognitive gap in recognizing disguised objects by leveraging existing prior knowledge. Notably, Grounding Dino \cite{dino} exhibits exceptional performance through its robust prior semantic knowledge.
 
 

While the aforementioned large models have demonstrated favorable outcomes, it is worth noting that they are not specifically designed for RCOD detection tasks, and prior knowledge is generally derived from well-defined contour scenes. They may exhibit domain bias and cognitive bias towards camouflaged samples. To address these limitations, we introduce CAFR, aimed at enhancing the model's capacity to differentiate between foreground and background, thereby improving the performance of large models, e.g., GLIP \cite{glipv2} and Grounding Dino \cite{dino} in the context of RCOD. As shown in Table \ref{datasets}, the CAFR technique consistently boosts the performance of these multimodal detection models, particularly concerning mAP and AP75 metrics. These findings compellingly illustrate the effectiveness of our approach in elevating the performance of COD tasks by refining the model's ability to distinguish between foreground and background.

\textbf{Results on NC4K-D.}
Similar to COD10K-D, we initially demonstrate the validation results of traditional detection models on the NC4K-D datasets. From Table \ref{datasets}, some models rely on a basic guessing strategy to detect camouflaged objects, e.g., YOLOv7 \cite{yolov7}. On the contrary, large detection models can achieve a rough estimation on camouflaged objects (with an mAP approaching 50\%, and even close to 80\% for Grounding Dino) on the NC4K-D dataset. Furthermore, our CAFR overcomes the bottleneck of large models by refining the model's ability to distinguish between foreground and background. For example, as shown in Table \ref{datasets}, there are 1.7\% performance improvements on GLIP-T, 0.5\% on GLIP-L, 0.7\% on GDino-T \cite{glipv2}, and 1.3\% on GDino-B \cite{dino}. Such results indicate that our CAFR can fully utilize a clear perception of the current object within the prior knowledge of large models to assist detectors in deeply understanding the distinctions between background and foreground.

\textbf{Results on CAMO-D.} As shown in Table \ref{datasets}, there also are significant performance breakthroughs in large detection models on the CAMO-D dataset, e.g., the GLIP-T model outperforms the FRCN+R101 by 23.3\%, and the GDino-T model outperforms the FRCN+R101 by 38.7\%. More importantly, our CAFR overcome the bottleneck of these large models. For example, as shown in Table \ref{datasets}, there are 0.6\% performance improvements on GLIP-T, 0.5\% on GLIP-L, 1.0\% on Grounding Dino-T, and 1.4\% on GDino-B on the CAMO-D dataset. Such results indicate that large models with CAFR can enhance the model's capacity to differentiate between foreground and background to improve the performance of object detection in camouflaged scenarios.

\begin{table}[t]
	\centering
	\tabcolsep=0.20cm
	\renewcommand{\arraystretch}{1.2}

	\caption{
		Ablation Study on AGP and SFR components.}
	
	\begin{tabular}{cc|ccc|ccc}
		\hline			
		\multicolumn{2}{c|}{Blocks}  & \multicolumn{3}{c|}{COD10K-D}   & \multicolumn{3}{c}{NC4K-D}                                                                                        \\

		APG   & SFR    & mAP & AP50 & AP75     & mAP & AP50 & AP75         \\  \hline
		
		\multicolumn{2}{c|}{Baseline} &  26.4  & 36.3  & 28.5   &  49.6  & 63.7  & 53.4 \\
		
		\CheckmarkBold &   & 27.8  & 38.3  & 29.9 &  50.9  & 65.2  & 53.7 \\
		&\CheckmarkBold  &  27.6  & 37.0  & 29.3 & 50.1 & 63.9 & 53.8\\
		\CheckmarkBold &\CheckmarkBold  & \textbf{28.8} & \textbf{38.2} & \textbf{31.0}  &  \textbf{51.3}& \textbf{66.7} & \textbf{53.9}                            \\

		\hline
	\end{tabular}	
	\label{AS}
	\vspace{-0.3cm} 
\end{table}

\subsection{Ablation Study on SFR and AGP}

Table \ref{AS} illustrates the impact of SFR and AGP on the performance of the model  GLIP-T \cite{glipv2} on the COD10K-D and NC4K-D datasets. We assess the contribution of the APG and SFR components by comparing different combinations of the mAP, AP50, and AP75 metrics. As can be seen from the table, APG and SFR components significantly enhance performance, with the combined inclusion of both SFR and AGP components yielding the best results. Specifically, on COD10K-D and NC4K-D datasets, mAP increased by 2.4\% and 1.7\%, AP50 by 1.9\% and 3.0\%, and AP75 by 2.5\% and 0.5\%, respectively. These findings indicate that reducing the size of the camouflaged object can optimize the efficiency of the transformer-based feature extractor and appropriately fine-tuning the gradient scale positively can fully refine class-specific features from camouflaged scenarios.

\begin{figure}[t]
	\begin{center}
		\includegraphics[width=0.45\textwidth]{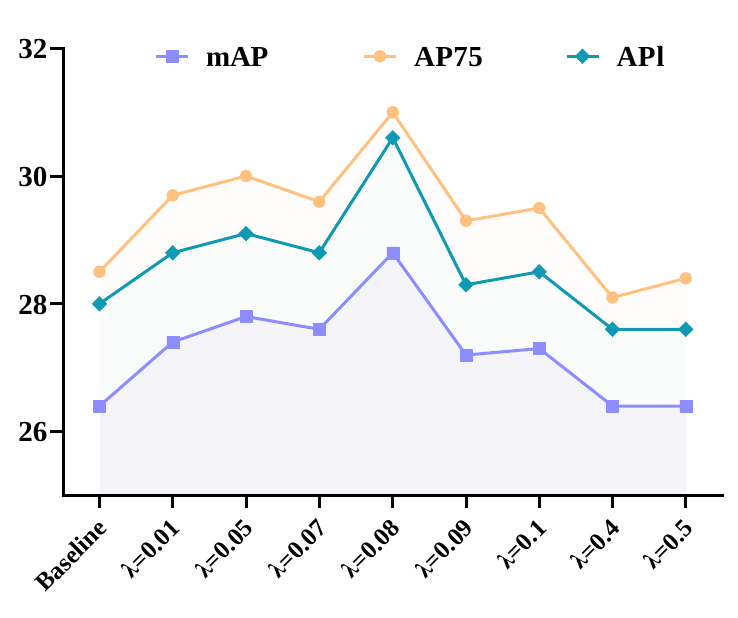}
	\end{center}
	\setlength{\abovecaptionskip}{-0.1cm} 
	\caption{Preformance of AGP in various parameter setting.}
	\vspace{-0.3cm} 
	\label{PC}
	\vspace{-0.2cm} 
\end{figure}

\subsection{Preformance of AGP in Various Parameter Settings}
To adapt and transfer rich pre-trained knowledge from large models to the RCOD task, we freeze the feature extractor of GLIP and focus solely on training the final fully connected (FC) layer, i.e., GLIP+FC. However, the results presented in Table \ref{adaptertable} reveal that training only this final FC layer does not enhance the generalization of large models for COD tasks. This finding suggests that fine-tuning the final layer parameters makes it hard to effectively capture these easily confused features since the foreground and background of camouflaged objects are extremely similar.

In addition, we introduce an adapter module composed of two 2-dimensional convolutional layers immediately after the feature extractor from the GLIP tiny model \cite{glipv2}. However, compared with the GLIP model, GLIP+Adapter fails to adapt to the COD task and forgets the prior knowledge of the large model, as shown in Table \ref{adaptertable}. This also proves that adding additional parameters to the last layer of the feature extractor and fine-tuning is not suitable for tasks where the background and foreground are extremely similar. On the other hand, we conjecture that discrepancy parameters and structural differences between GLIP and the GLIP with adapter network impede the direct adaptation of pre-trained model weights to the new network configuration. This could result in the model getting stuck in a local optimum or failing to learn task-specific features correctly. Consequently, the model may have become trapped in a suboptimal solution or struggled to learn task-specific features accurately.

To address the mentioned issue, we propose the AGP. The results in Table \ref{AS} demonstrate that AGP effectively fine-tunes the prior knowledge of the large model for the RCOD task. Moreover, we conduct a comparative experiment of parameter  $\lambda$ to find the optimal solution in AGP, as shown in Fig. \ref{PC}. From the figure, it can be observed that the model achieves the optimal solution in the COD10K-D dataset when  $\lambda$ is set to 0.08. 
Furthermore, the average values of all the results presented in the figure exceed the baseline, suggesting that AGP incorporates only a limited number of parameter updates across all feature extraction layers. This approach effectively optimizes the RCOD task while preventing the introduction of additional parameters that could impede the model's ability to adapt its weights to the network directly.


\subsection{Preformance of SFR in Various Parameter  Settings}	
\begin{figure}[t]
	\begin{center}
		\includegraphics[width=0.45\textwidth]{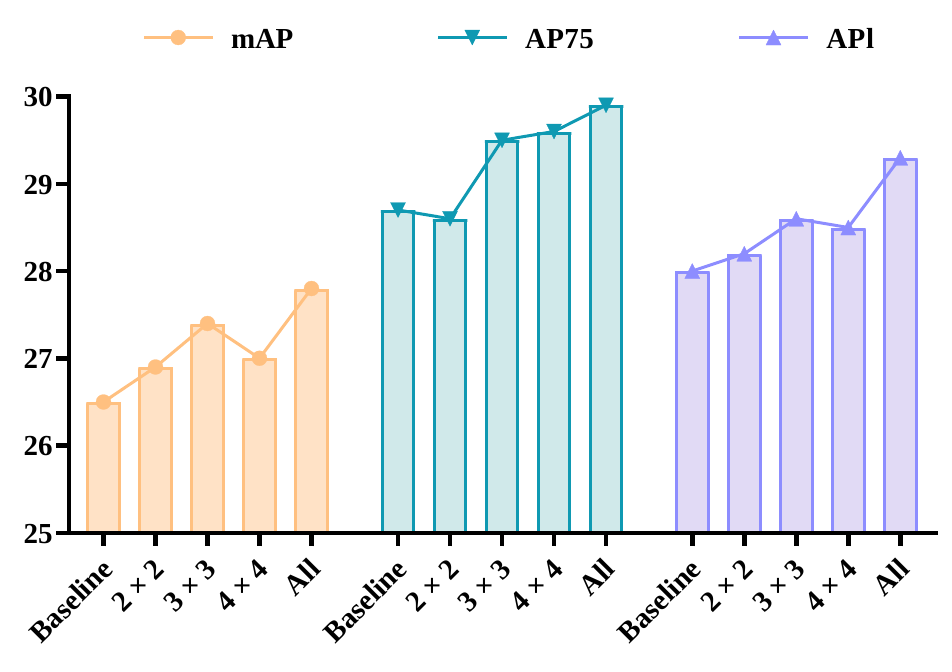}
	\end{center}
	\setlength{\abovecaptionskip}{-0.1cm} 
	\caption{Preformance of SFR in parameter various setting. All indicates the result of combining $2 \times 2$, $3 \times 3$, and $4 \times 4$.}
	\vspace{-0.3cm} 
	\label{SR}
	\vspace{-0.2cm} 
\end{figure}

To verify the optimal performance of SFR, Fig. \ref{SR} presents the improvement results of different bounding box concatenations, including $2 \times 2$, $3 \times 3$, and $4 \times 4$, based on the GLIP-T model \cite{glipv2} on the COD10K-D dataset. In an input batch, each box is cropped to $W=200$, $H=200$, and randomly placed on a square canvas. From the table, compared to the baseline, we find that using $2 \times 2$, $3 \times 3$, and $4 \times 4$ can respectively enhance the performance of the GLIP-T model, especially in terms of AP75. It is worth noting that the result of combining $2 \times 2$, $3 \times 3$, and $4 \times 4$ achieved the best performance.  In addition, we explore the impact of combining these concatenations in a batch. By training the GLIP-T model with a mixture of all concatenations, the greatest performance improvement is achieved, i.e., 2.3\% improvement in mAP over the baseline.
Such results illustrate that SFR alleviates the issue of feature confusion by using transformer-based feature extractors when dealing with large-scale camouflaged objects with sparse features.

\begin{table}[t]
	\centering	
	\tabcolsep=0.20cm
	\renewcommand{\arraystretch}{1.2}	
	\caption{Performance comparison between online and offline of SFR. We use GLIP-T pre-trained model \cite{glipv2} to fine-tune the COD task on the COD10K-D dataset.}
	\begin{tabular}{l|c|cccc}
		\midrule

		method & Cuda Memory (G) & mAP  & AP75  &  APl & Avg\\ \midrule		
		
		SFR-Online & 22.27  & 27.8  & 38.3  & 29.9 & 32.0 \\

		SFR-Offline &  20.95 & 27.8  & 38.0  & 30.1 & 32.0\\

		\bottomrule
		
	\end{tabular}

	\label{srof}
	
\end{table}

\begin{table}[t]
	\centering	
	\tabcolsep=0.25cm
	\renewcommand{\arraystretch}{1.2}	
	\caption{Influence of parameter fine-tuning on the large model in the COD task.}
	\begin{tabular}{l|ccccc}
		\midrule

		method & mAP  & AP75  &  APl & Avg\\ \midrule		
		
		GLIP &  26.4   & 28.5    & 28.0 & 27.6 \\

		GLIP + FC  & 26.4  & 28.4   & 28.6  & 27.8 \\		
		
		GLIP + Adapter  &25.7  &27.9  &26.6 & 26.7\\
		
		GLIP + AGP &  \textbf{27.4}    & \textbf{29.7}    & \textbf{28.8} & \textbf{28.6} \\

		\bottomrule
		
	\end{tabular}

	\label{adaptertable}
	
\end{table}

\begin{figure*}[t]
	\begin{center}
		\includegraphics[width=1\textwidth]{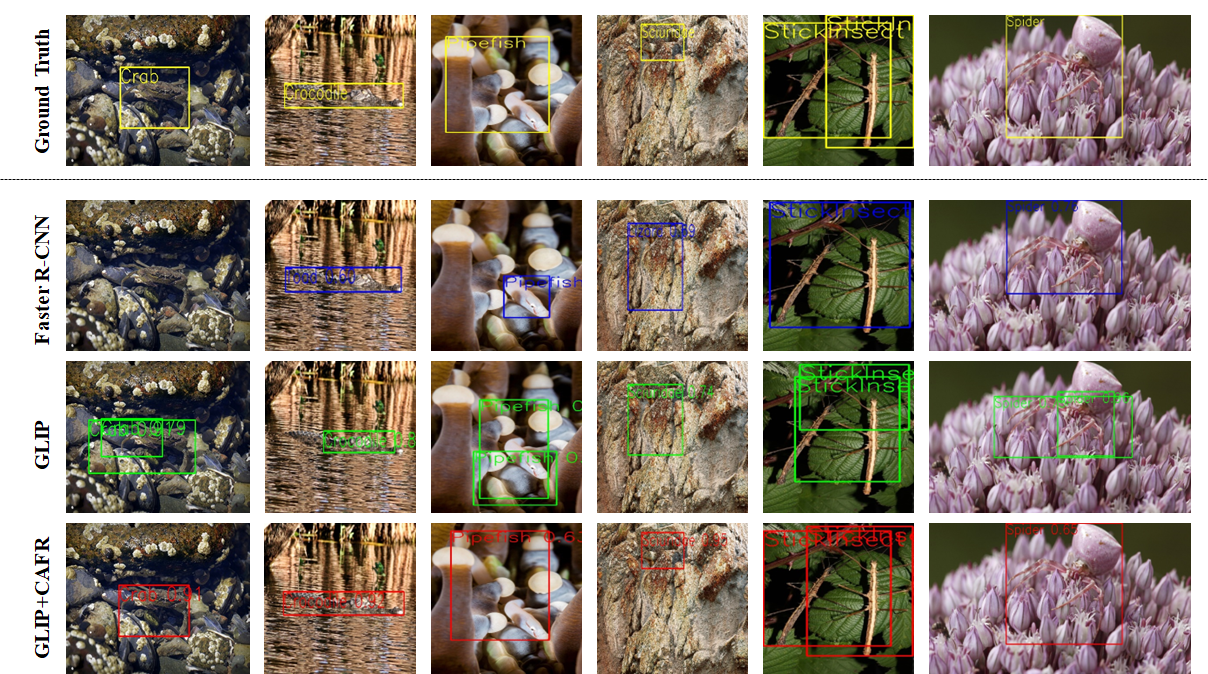} 
	\end{center}
	\setlength{\abovecaptionskip}{-0.1cm} 
	\caption{Visualization in Camouflaged Scenarios.}
	\vspace{-0.3cm} 
	\label{vis}
	\vspace{-0.2cm} 
\end{figure*}
\subsection{Evaluation of two SFR Implementations}
\label{online}
Our SFR method consists of two processing approaches: online and offline. The online approach offers the advantage of directly processing input data to obtain scaled-down data, providing greater flexibility and convenience. However, it necessitates additional GPU resources. In contrast, the offline approach eliminates the need for extra CUDA memory usage but requires manual pre-processing of data, making it less flexible compared to the online method. To evaluate the performance of these two approaches, we train the GLIP-T model on a single GPU (RTX 3090Ti) with a batch size of 6. The experimental results, as shown in Table \ref{srof}, indicate that, after multiple rounds of averaging, the accuracy differences between the two methods are minimal. Consequently, the selection between the online and offline approaches can be tailored effectively based on the specific requirements of the RCOD task.

\subsection{Visualization in Camouflaged Scenarios}

We visualize a comparison of three object detection methods, i.e., Faster R-CNN \cite{fasterrcnn}, GLIP \cite{glipv2}, and GLIP-CAFR, in camouflaged scenarios using the COD10K-D dataset (refer to Fig. \ref{vis}). In the case of the Faster R-CNN detector, we utilize swin-transformer as the feature extractor. Although Faster R-CNN demonstrates roughness in object localization, it encounters difficulties in accurately categorizing the objects, relying heavily on guesswork. Conversely, GLIP exhibits precise classification capabilities, however, it suffers from problems associated with over-scaling and insufficient generalization, resulting in numerous redundant detections. In contrast, our GLIP-CAFR method outperforms the other approaches in the tested scenarios by providing higher accuracy and robustness.
 Such visualization demonstrates that the CAFR strategy can assist the large detection model to fully refine class-specific features from camouflaged contexts and enhance the efficiency of the self-attention mechanism for the RCOD task.

\section{Conclusion and Discussion}

In this paper, we have proposed an CAFR strategy for object detection in camouflaged scenarios and introduced three novel datasets as new benchmarks for evaluating the performance of the RCOD task. In the CAFR strategy, we have designed a AGP module to fully refine class-specific features from camouflaged contexts and an SFR module to optimize the performance of transformer-based feature extractors in RCOD tasks. Extensive experiments conducted on our proposed datasets have demonstrated that the CAFR strategy can enhance the performance of the detection model for the RCOD task.

In object detection tasks, particularly those involving datasets with class imbalances, traditional object detection methods encounter ongoing research challenges. This is especially true in scenarios like RCOD, where backgrounds closely resemble foregrounds. Fine-tuning large detection models for quick adaptation to RCOD presents a promising approach for rapidly improving detection accuracy, though it may not be the most optimal solution. We are confident that future efforts will continue to advance toward achieving faster speeds and higher accuracies.

\bibliographystyle{IEEEtran}
\bibliography{aaai25}

\end{document}